\documentclass[10pt,journal,compsoc,twoside]{IEEEtran}
\usepackage{amsmath,amsfonts}
\usepackage{algorithmic}
\usepackage{algorithm}
\usepackage{array}
\usepackage[caption=false,font=normalsize,labelfont=sf,textfont=sf]{subfig}
\usepackage{textcomp}
\usepackage{stfloats}
\usepackage{url}
\usepackage{verbatim}
\usepackage{graphicx}
\usepackage{cite}
\usepackage{siunitx}
\usepackage{makecell}
\usepackage{caption}
\usepackage{booktabs}
\newcommand{\ra}[1]{\renewcommand{\arraystretch}{#1}}

\usepackage{amsmath,amssymb}
\newcommand{\R}{{\mathbb{R}}}
\usepackage{color}
\definecolor{darkorange}{rgb}{1.0, 0.55, 0.0}
\definecolor{blue}{rgb}{0.0, 0.0, 1.0}
\definecolor{nicegreen}{rgb}{0.0, 0.7, 0.1}

\usepackage{tikz}
\usepackage{hyperref}
\usepackage{multirow}

\makeatletter
\long\def\@IEEEtitleabstractindextextbox#1{%
    \begin{center}
        \captionsetup{type=figure}
        \setcounter{figure}{0}
        \begin{tikzpicture}
            \newcommand\smallermath{\fontsize{6pt}{7pt}\selectfont}
            \node (img) {\includegraphics[width=0.99\textwidth]{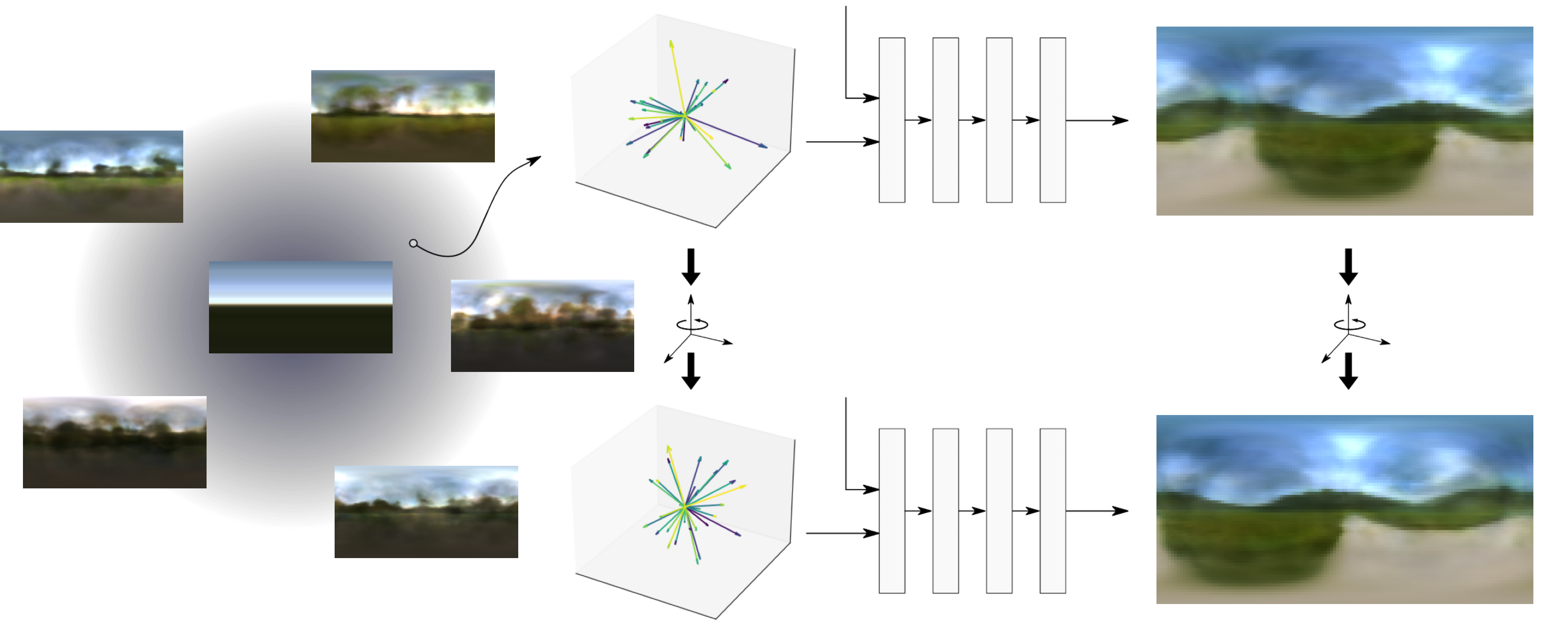}};
            \node at (-5.5, -0.65){\small $\mathbf{Z}=\mathbf{0}_{3\times N}$};
            \node at (-5.0, 1.0) {\small $\text{vec}(\mathbf{Z})\sim\mathcal{N}(\mathbf{0},\mathbf{I}_{3N})$};
            \node at (0.65, 1.8) {\small $\mathbf{Z}$};
            \node at (0.65, -2.7) {\small $\mathbf{Z}$};
            \node at (0.71, 3.70) {\small $\mathbf{D}$};
            \node at (0.71, -0.80) {\small $\mathbf{D}$};
            \node at (2.8, 0.2) {\footnotesize SO(2) Rotation of Latent Code};
            \node at (2.8, -0.3){\footnotesize Around Vertical Y-Axis};
            \node at (-1.65, -3.4) {\smallermath $x$};
            \node at (-0.1, -3.1) {\smallermath $z$};
            \node at (0.25, -2.1) {\smallermath $y$};
            \node at (-1.65, 1.1) {\smallermath $x$};
            \node at (-0.1, 1.4) {\smallermath $z$};
            \node at (0.25, 2.4) {\smallermath $y$};
        \end{tikzpicture}
        \captionof{figure}{On the left, we visualize environment maps derived from random latent samples of RENI++, our natural illumination prior, as well as the average illumination in the centre. RENI++ is rotation-equivariant to rotations of the latent codes around the vertical \(y\)-axis (right). Plots show a latent code at two rotations and the resulting output of the RENI++ network equally rotated.}
        \label{fig:teaser}
    \end{center}
    \parbox{0.922\textwidth}{#1}
}
\makeatother
\begin{document}

\title{RENI++: A Rotation-Equivariant, \\ Scale-Invariant, Natural Illumination Prior}

\author{James~A.~D.~Gardner,
        Bernhard~Egger,
        and~William~A.~P.~Smith,% <-this % stops a space
\IEEEcompsocitemizethanks{\IEEEcompsocthanksitem J. A. D. Gardner and W. A. P. Smith are with the Department of Computer Science, University of York, York, United Kingdom. \protect\\
E-mail: \{james.gardner, william.smith\}@york.ac.uk
\IEEEcompsocthanksitem B. Egger is with the Cognitive Computer Vision Lab, Friedrich-Alexander-Universität Erlangen-Nürnberg, Erlangen, Germany. \protect\\
E-mail: bernhard.egger@fau.de}% <-this % stops a space
}

% \markboth{}%
\markboth{}%
{Gardner \MakeLowercase{\textit{et al.}} RENI++ A Rotation-Equivariant, Scale-Invariant, Natural Illumination Prior}

\IEEEtitleabstractindextext{%
\begin{abstract}
Inverse rendering is an ill-posed problem. Previous work has sought to resolve this by focussing on priors for object or scene shape or appearance. In this work, we instead focus on a prior for natural illuminations. Current methods rely on spherical harmonic lighting or other generic representations and, at best, a simplistic prior on the parameters. This results in limitations for the inverse setting in terms of the expressivity of the illumination conditions, especially when taking specular reflections into account. We propose a conditional neural field representation based on a variational auto-decoder and a transformer decoder. We extend Vector Neurons to build equivariance directly into our architecture, and leveraging insights from depth estimation through a scale-invariant loss function, we enable the accurate representation of High Dynamic Range (HDR) images. The result is a compact, rotation-equivariant HDR neural illumination model capable of capturing complex, high-frequency features in natural environment maps. Training our model on a curated dataset of 1.6K HDR environment maps of natural scenes, we compare it against traditional representations, demonstrate its applicability for an inverse rendering task and show environment map completion from partial observations. We share our PyTorch implementation, dataset and trained models at \href{https://github.com/JADGardner/ns_reni}{github.com/JADGardner/ns\_reni}.
\end{abstract}

}

\IEEEpeerreviewmaketitle
\maketitle

\section{Introduction}

\IEEEPARstart{T}{he} human visual system is capable of solving under-constrained inverse rendering problems. Many different combinations of shape, materials and lighting can produce the same observed image \cite{belhumeur_bas-relief_1999}. Hence, in order to estimate the shape, material properties and colour of surfaces and objects in a scene, the human visual system draws on priors that seek the most likely explanation. These include strong priors over the space of possible illuminations \cite{murray2019visual}. For example, there is wide evidence that humans exploit a lighting-from-above or increasing-luminance-with-elevation prior \cite{hill_independent_1993, thomas_interactions_2010}. It has also been shown that humans perform inverse rendering tasks better under complex natural illumination while performance degrades when the illumination statistics are not representative of the real world \cite{fleming2003real}. Interestingly, it seems that these illumination priors are, at least partially, learnt \cite{granrud1985infants,thomas2010interactions} and can be updated from experience \cite{adams2004experience}.

Compact but expressive lighting representations play an essential role in graphics, enabling realistic lighting effects at real-time frame rates \cite{ramamoorthi_efficient_2001, tsai_all-frequency_2006, wang_all-frequency_2009, ng_all-frequency_2003, green_spherical_2003} and in computer vision, enabling scene relighting \cite{yu_outdoor_2021, yu_self-supervised_2020}, face relighting \cite{wang_face_2009, shu_neural_2017, egger_occlusion-aware_2018, sengupta_sfsnet_2018} and object insertion \cite{wang_learning_2021, song_neural_2019, li_inverse_2020}. Real-world illumination is highly complex and variable, with a very high dynamic range, and is therefore inherently challenging to represent. However, real-world illumination does contain statistical regularities \cite{dror_statistical_2004}, particularly for outdoor, naturally lit scenes. The lighting-from-above prior holds in a world where the strongest illumination source is the sun or skylight, which also produce only a limited range of colours. In addition, illumination environments have a canonical up direction (vertical axis aligns with gravity) but arbitrary horizontal rotation (any rotation about the vertical is equally likely). We also desire vision systems to have exposure invariance (i.e.~a human would infer the same inverse rendering result regardless of pupil dilation) and so an illumination environment could be encountered with any absolute scale. These regularities and geometric symmetries can significantly restrict the space of possible illuminations to constrain inverse problems or enable the synthesis of realistic lighting.

Given this, it is surprising that statistical illumination priors have been almost completely ignored in computer vision, with the vast majority of inverse rendering techniques allowing arbitrary illumination within their chosen representation space. In this paper we attempt to replicate the sort of illumination prior used by the human visual system. This entails learning a statistical characterisation of the illumination environments likely to be encountered in the real world but also incorporating invariances and symmetries by construction. Specifically, we desire a representation of natural illumination environments that exhibits the following features:
\begin{itemize}
    \item \emph{Generative:} A generative model that captures the statistical regularities of natural illumination with a well-behaved latent space within which we can optimise to solve inverse problems.
    \item \emph{Compact:} Reduces the dimensionality of inverse problems while preserving high-frequency lighting effects that are important for non-Lambertian appearance.
    \item \emph{Rotation-Equivariant:} Respect the canonical orientation, i.e.~any rotation of an environment about the vertical should be equally likely and equally well represented.
    \item \emph{Scale-Invariant:} Any scaling of the exposure of an environment should be equally likely and equally well represented.
    \item \emph{Statistical Prior:} Provides a prior to regularise inverse problems, or that can be sampled from for synthesis, only generating plausible illumination environments.
    \item \emph{HDR:} Correctly handle HDR quantities essential for realistic rendering and the reproduction of natural light.
\end{itemize}

\subsection{Contributions}

We introduce RENI++ - A Rotation-Equivariant Natural Illumination model. In so doing, we make the following key contributions:
\begin{itemize}
\item[--] An extension of Vector Neurons to a rotation-equivariant neural field representation for spherical images, optionally restricted to rotations about the vertical axis.
\item[--] A variational autodecoder architecture for a generative model of spherical signals.
\item[--] The first natural, outdoor HDR illumination model.
\item[--] Evaluated in an inverse rendering task showing significant performance improvements over other lighting representations.
\end{itemize}
We choose to model scene radiance, i.e.~environment lighting, directly as opposed to pre-integrated lighting with a particular BRDF. This makes our model more general since it can be used with arbitrary BRDFs at inference time or even for tasks other than rendering, such as to constrain shape from specular flow.

This work is an extension of our earlier RENI model \cite{gardner2022rotation}. Compared to our earlier work, RENI++ adds: 
\begin{itemize}
\item[--] A scale-free loss enabling a large increase in generalisation capabilities on previously out-of-distribution HDR environments.
\item[--] Transformer-decoder architecture with positional encoding, replacing the previous SIREN network.
\item[--] An invariant representation that scales as \(O(n)\) rather than \(O(n^{2})\) with the size of latent space by replacing the Gram-Matrix with the VN-Invariant layer \cite{deng_vector_2021}.
\item[--] A new NeRFStudio-based \cite{nerfstudio} implementation with an order-of-magnitude speedup in training time. 
\item[--] A new data normalisation, augmentation and sampling procedure.
\end{itemize}
\subsection{Overview}

In Section \ref{sec:related_work} we review prior work in the areas of lighting representations, illumination priors, neural fields and invariances. In Section \ref{sec:RECSNF} we introduce our rotation-equivariant, conditional neural field representation for spherical signals. In Section \ref{sec:RENI} we describe how this framework is applied to the task of learning a natural illumination prior, including the use of a scale-invariant loss. In Section \ref{sec:eval} we evaluate our model for generalisation, interpolation, completion, inverse rendering and LDR to HDR. We also perform an ablation study of our design decisions.

\section{Related Work}\label{sec:related_work}

\subsection{Lighting Representations} 

\begin{figure*}
    \centering
    \makebox[\textwidth]{%
        \begin{tikzpicture}
    
        \node (img) {\includegraphics[width=0.98\textwidth]{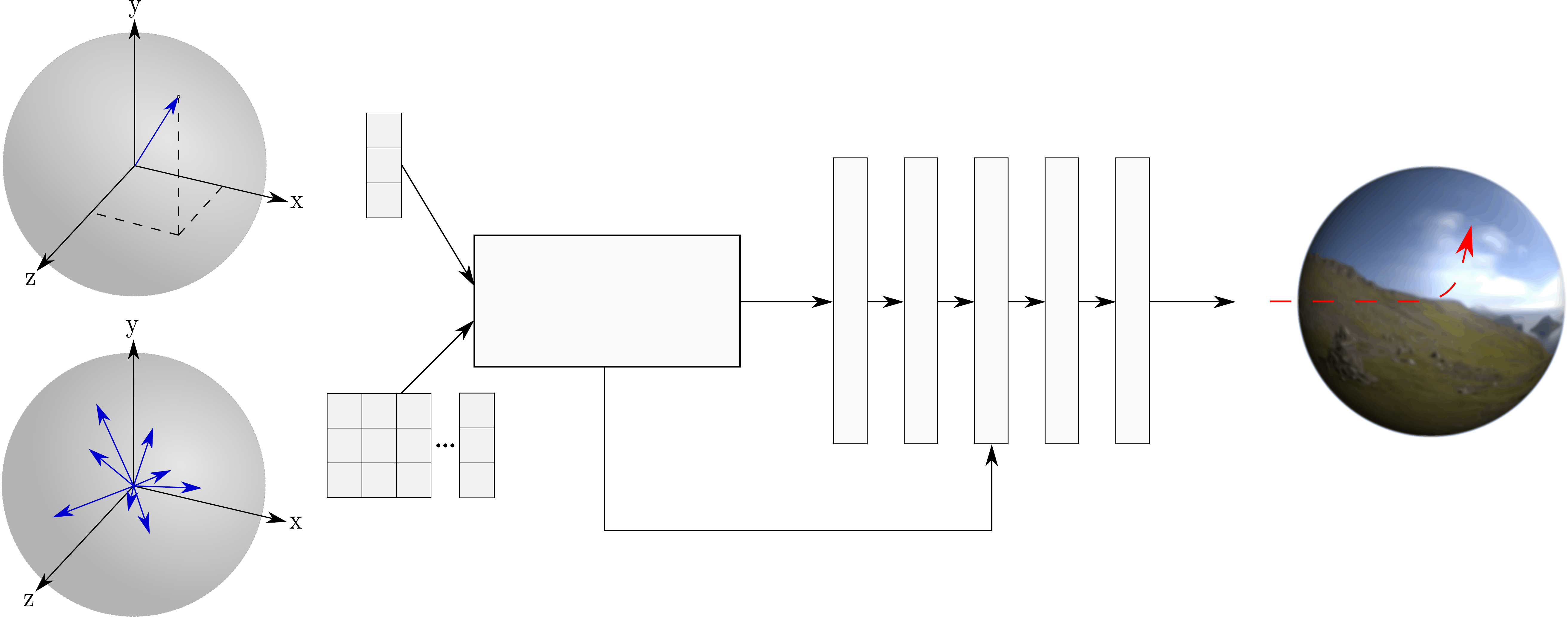}};
        \node at (-4.56,2.04) {$x$};
        \node at (-4.56,1.63) {$y$};
        \node at (-4.56,1.24) {$z$};
        \node at (-4.25,-2.5) {$3 \times N$};
        \node at (-6.7,2.5) {$\mathbf{d}$};
        \node at (-6.7,-1.5) {$\mathbf{Z}$};
        \node at (0.0,-0.2) {$\mathbf{d}^\prime$};
        \node at (0.25,-2.2) {$\mathbf{Z}^\prime$};
        \node at (5.3, 0.08) {$\mathbf{C}$};
        \node at (2.35,2.5) {Rotation-Equivariant Conditional};
        \node at (2.35,2.1){Spherical Neural Field};
        \node at (-2.0,0.3){Invariant}; 
        \node at (-2.0,-0.1){Transformation};
        \node at (-8.5,2.9) [rotate=45] {\small Query};
        \node at (-8.5,-0.7) [rotate=45] {\small Latent Code};
        \end{tikzpicture}
    }
    \caption{We propose to represent a space of spherical signals via a rotation-equivariant conditional spherical neural field. The signal in a direction $\mathbf{d}$ can be queried by evaluation of the network and rotating the Vector Neuron conditioning latent code $\mathbf{Z}$, corresponds to rotating the spherical signal.}
\end{figure*}

An illumination environment is a spherical signal. A relatively small set of alternatives are used for their representation within vision and graphics. A widely used representation in graphics is an environment map \cite{ramamoorthi_frequency_2002, ramamoorthi_efficient_2001, tsai_all-frequency_2006, wang_all-frequency_2009}, which is a regularly sampled 2D image representing a flattening of the sphere, usually via an equirectangular projection. However, the projection introduces distortions leading to irregular sampling on the sphere, it is not compact, introduces boundaries and provides no prior for inverse problems. Nevertheless, environment map representations have been used in inverse settings where every pixel in the map is optimised independently \cite{sengupta_neural_2019}.

Spherical harmonic (SH) lighting \cite{basri_lambertian_2003,ramamoorthi_efficient_2001} is a compact lighting representation commonly used in real-time computer graphics \cite{ramamoorthi_efficient_2001, green_spherical_2003, sloan_precomputed_2002} and inverse rendering \cite{yu_outdoor_2021, tsai_all-frequency_2006, li_inverse_2020, rudnev_nerf_2022, egger_occlusion-aware_2018}. While SHs can be used to represent the illumination environment directly, more commonly, they represent pre-integrated lighting, i.e.~the illumination environment convolved with a bidirectional reflectance distribution function (BRDF). When the BRDF is low frequency, as it is for Lambertian diffuse reflectance, then the convolution is also low frequency making the approximation with SHs very accurate \cite{basri_lambertian_2003}. 

An alternative, growing in popularity, is the Spherical Gaussian (SG) representation \cite{wang_all-frequency_2009, tsai_all-frequency_2006, zhang_physg_2021}. SGs represent a lighting environment as a collection of Gaussian lobes on the sphere, each of which has 6 degrees of freedom (three for RGB amplitude, two for spherical direction and one for sharpness). While this allows the reconstruction of localised high-frequency features, it still requires many lobes to approximate complex illumination environments. \cite{li_inverse_2020} compares SH and SG for object re-lighting, finding SG was able to recover higher frequency lighting using a similar number of parameters as SH, though both still required a large number of parameters to approximate ground truth.

Both SHs and SGs are \emph{rotation equivariant}. A rotation of the illumination environment corresponds directly to a rotation of the SH basis or the SG lobe directions. Equivalently, they can represent any rotation of a given environment with equal accuracy. However, they provide no prior over the space of possible illuminations. SGs or SHs can represent any colour of light coming from any direction.

\subsection{Illumination Priors} 

While it has long been known that natural illumination environments exhibit statistical regularities \cite{dror_statistical_2004}, this has largely been overlooked in prior work. Here we describe the small number of exceptions. Barron and Malik \cite{barron2014shape}, Egger et al.~\cite{egger_occlusion-aware_2018}, and Yu and Smith \cite{yu_outdoor_2021} learn a linear statistical model with Gaussian prior in the space of SH coefficients. Both Barron and Malik \cite{barron2014shape} and Yu and Smith \cite{yu_outdoor_2021} learn their model from a curated set of known environments maps, while Egger et al.~\cite{egger_occlusion-aware_2018} use indirectly observed illumination environments estimated from fitting a morphable model to face images. Barron and Malik \cite{barron2014shape} build their model in log space as they found the Gaussian assumption on SH parameters to hold better. While providing a useful constraint to avoid unrealistic illumination environments, these approaches inherit the weakness of SHs in being unable to reproduce high-frequency lighting effects while also losing the rotation equivariance. Yu and Smith \cite{yu_outdoor_2021} seek to overcome this by rotation augmentation at training time, but this brute force approach makes no guarantee of rotation equivariance. 

Sztrajman et al.~\cite{sztrajman_high-dynamic-range_2020} separate environment maps into HDR and LDR components. Using a CNN-based auto-encoder for estimations of LDR components of lighting alongside a low dimensional SG model for the HDR lighting provided by the sun. They too require data augmentation in the form of rotations, can make no equivariance guarantees and, due to the low dimensionality of the SG model, struggle to represent environments with multiple HDR light sources.

Several other neural-based HDR environment map models have followed since RENI. Both Somanath et al.~\cite{Somanath_2021_CVPR} and Dastjerdi et al.~\cite{dastjerdiEverLightIndoorOutdoorEditable2023} learn to predict a full HDR environment map from a narrow field-of-view LDR camera image. Unlike our work, Somanath et al.~\cite{Somanath_2021_CVPR} do not learn a prior and cannot sample from their latent space. Dastjerdi et al.~\cite{dastjerdiEverLightIndoorOutdoorEditable2023} allows parametric control over the lighting direction a Spherical Gaussian representation of the dominant light source. Chen at el.~\cite{chenText2LightZeroShotTextDriven2022} learn a text-to-HDR environment map model via a CLIP \cite{radfordLearningTransferableVisual2021} conditioned global code-book sampler and a structure-aware local code-book sampler. They propose a novel Super-Resolution Inverse Tone Mapping Operator (SR-iTMO) to simultaneously increase the spatial resolution
and dynamic range of the environment map. Yao et al.~\cite{yaoNeILFNeuralIncident2022} model incident light at a point for a given direction as a 5D light field network allowing the modelling of spatially varying illumination effects. Lyu et al.~\cite{lyuDiffusionPosteriorIllumination2023a} train an unconditional diffusion model \cite{hoDenoisingDiffusionProbabilistic2020} to generate realistic environment maps. Then, in an inverse rendering setting, gradients from a differentiable path tracer are used to update the posterior score function such that the diffusion model generates an environment map to explain the image.

\subsection{Neural Fields}

Neural fields \cite{xie_neural_2021} have provided impressive results in a range of applications including representations of objects and scenes \cite{sitzmann_implicit_2020, atzmon_sal_2020, mescheder_occupancy_2019, chibane_neural_2020, chen_learning_2019, gropp_implicit_2020, park_deepsdf_2019, tancik_block-nerf_2022}, in inverse rendering \cite{boss_nerd_2021, bi_neural_2020, rudnev_nerf_2022, srinivasan_nerv_2020, boss_neural-pil_2021, zhang_physg_2021} and robotics \cite{li_3d_2021, chen_full-body_2021, ortiz_isdf_2022}. The work most closely related to ours is Neural-PIL \cite{boss_neural-pil_2021}. They use a FiLM SIREN similar to that proposed in Pi-GAN \cite{chan_pi-gan_2021}, and like us use a direction query vector and auto-decoder architecture. However, they do not have rotation-equivariance and apply no natural light prior. They also model pre-integrated lighting, conditioning the latter layers of their FiLM SIREN on a material roughness parameter. \cite{rebainAttentionBeatsConcatenation2023} evaluated the effectiveness of three possible conditioning methods for neural fields and found that attention-based conditioning was the most effective. In RENI++ we too found that a transformer-based architecture outperformed both a conditioned-via-concatenation and FiLM conditioned SIREN.

\subsection{Invariance and Equivariance}

Two important symmetries in computer vision are invariance and equivariance to the rotation group \cite{bronstein_geometric_2021}. Some works attempt to achieve these properties via data augmentation \cite{yu_outdoor_2021, qi_pointnet_2017, sztrajman_high-dynamic-range_2020}, which still results in missed cases for continuous rotations. A solution alleviating the need for extensive data augmentation is the Vector Neuron \cite{deng_vector_2021}, which offers a framework for designing SO(3)-Equivariant networks via a latent matrix representation rather than a latent vector. This allows a direct mapping of rotations applied to the network's input to its output, resulting in all possible rotations being explicitly represented via rotations of the latent codes. Vector Neurons were originally designed for rotation equivariant or invariant processing of point cloud data. We re-purpose them for the representation of spherical signals such as environment maps. Another commonly sought invariance is to scale. For example, in the monocular depth estimation problem, absolute depth is often unobtainable due to a lack of calibration information. For this reason, scale-invariant losses \cite{eigenDepthMapPrediction2014} are often used, such that the depth prediction network learns to estimate only relative scale. The output depth map is then only valid up to an unknown scale. We take the same approach in learning a model of natural illumination. An environment map is a representation of a real environment but at arbitrary scale (dependent on camera exposure and the subsequent image processing pipeline). Our model therefore learns the relative brightness of natural environments and the absolute scale becomes an additional optimsable parameter when fitting to data.
\section{Rotation-Equivariant Conditional Spherical Neural Fields}\label{sec:RECSNF}

We wish to construct a generative model of spherical signals that is rotation equivariant with respect to the latent representation of the signal. That is to say, a rotation of the latent representation corresponds to a rotation of a spherical signal and a signal can be reconstructed with exactly the same accuracy in any rotation. We propose two variants of Vector Neurons \cite{deng_vector_2021} for $SO(2)$ and $SO(3)$ equivariant representation of spherical signals.

\subsection{Spherical signals as vector neurons}
As in Vector Neurons \cite{deng_vector_2021}, we use an ordered list of 3D vectors for our latent representation. In contrast to Vector Neurons \cite{deng_vector_2021}, our signal is defined over directions, not positions. 

Our model takes the form of a conditional spherical neural field, $f:S^2\times\R^{3\times N}\rightarrow \R^M$, such that $f(\mathbf{d},\mathbf{Z})$ computes the value of the signal represented by vector neuron latent code $\mathbf{Z}\in\R^{3\times N}$ in direction $\mathbf{d}\in\mathbb{R}^3$, with $\|\mathbf{d}\|=1$. Our rotation equivariant construction is independent of the specific architecture or conditioning mechanism of $f$. For colour images, $M=3$ and $f$ outputs an RGB colour. By using a spherical neural field, we are agnostic to how the signals are sampled on the sphere. We can generate any sampling simply by choosing the grid of directions as appropriate. We also avoid boundary effects since our domain is continuous. This is a significant advantage over methods for learning spherical signals that operate in the 2D image domain, for example using equirectangular projections.

\subsection{SO(3) Equivariance}

We construct the neural field such that it is \emph{invariant} to a rotation of both $\mathbf{d}$ and $\mathbf{Z}$ simultaneously (i.e.~$f(\mathbf{R}\mathbf{d},\mathbf{R}\mathbf{Z})=f(\mathbf{d},\mathbf{Z})$ with $\mathbf{R}\in SO(3)$). This entails that the neural field is \emph{equivariant} with respect to a rotation of $\mathbf{Z}$ only (i.e.~rotating $\mathbf{Z}$ corresponds to rotating the spherical signal such that $f(\mathbf{d},\mathbf{R}\mathbf{Z})=f(\mathbf{R}^\top\mathbf{d},\mathbf{Z})$). See a visualisation of this property on the right of Figure \ref{fig:teaser}.

Key to our approach is a transformation of the inputs to the neural field, $(\mathbf{d},\mathbf{Z})$, such that they are rotation invariant. These divide into two parts:
\begin{enumerate}
    \item $\mathbf{d}^\prime$ - the directional input to the spherical neural field,
    \item $\mathbf{Z}^\prime$ - the latent code on which the neural field is conditioned.
\end{enumerate}

So long as these two inputs satisfy the desired rotation invariance then the neural field itself exhibits this invariance.

The direction in which we wish to evaluate the spherical neural field must be encoded relative to the latent code in the particular rotation in which we encounter it. This is satisfied by using the inner product $\langle \mathbf{d},\mathbf{Z} \rangle$, i.e.~the matrix-vector product $\mathbf{d}^\prime=\mathbf{Z}^\top\mathbf{d}\in\R^N$. Unlike Vector Neurons, our input is a direction with unit norm, not a position. Hence, the rotation invariant feature $\|\mathbf{d}\|$ conveys no information and we do not use it.

For the latent code, we apply the Vector Neurons \cite{deng_vector_2021} invariant layer to the latent code: $\mathbf{Z}^\prime=\text{VN-In}(\mathbf{Z})$. Since our neural field is equivariant \emph{we do not need to augment our training data over the space of rotations}. Observing a spherical signal once in a particular rotation means we can reconstruct it with the same accuracy in any rotation by rotating the latent code.

\subsection{SO(2) Equivariance}\label{sec:SO2}

In certain settings it is desirable to restrict the equivariance to a specific subset of rotations. We propose a restricted transformation of the neural field inputs that are invariant only to rotations, $\mathbf{R}_{\mathbf{a}}\in SO(3)_{\mathbf{a}}\cong SO(2)$, about a given axis $\mathbf{a}\in\mathbb{R}^3$, $\|\mathbf{a}\|=1$, i.e.~a subgroup of $SO(3)$ that is isomorphic to $SO(2)$.

In order to construct the invariant features, we define $\mathbf{b}_{\perp\mathbf{a}}$ as the vector rejection of $\mathbf{b}$ onto $\mathbf{a}$. i.e.~the orthogonal projection of $\mathbf{b}$ onto the plane that is orthogonal to $\mathbf{a}$. Note that this component of $\mathbf{b}$ \emph{is} affected by $\mathbf{a}$-axis rotation. We can write the vector rejection of $\mathbf{b}$ onto $\mathbf{a}$ as a matrix multiplication by first rotating $\mathbf{a}$ onto an arbitrarily chosen standard basis, we choose $\mathbf{e}_x=[1,0,0]^\top$, followed by orthogonal projection and rotation back:
\begin{equation}
    \mathbf{b}_{\perp\mathbf{a}} = \mathbf{R}_{\mathbf{e}_x\rightarrow\mathbf{a}}\text{diag}(0,1,1)\mathbf{R}_{\mathbf{a}\rightarrow\mathbf{e}_x}\mathbf{b},
\end{equation}
where $\text{diag}(0,1,1)$ is the orthogonal projection onto the $y$-$z$ plane and $\mathbf{R}_{\mathbf{a}\rightarrow\mathbf{e}_x}$ is the rotation matrix that rotates $\mathbf{a}$ onto $\mathbf{e}_x$. Since we are not interested in the specific coordinate frame of the invariant features, we simplify by dropping the second rotation and only retaining the $y$ and $z$ coordinates (the $x$ coordinate will be zero in the rotated space). This provides 2D coordinates:
\begin{equation}
    \mathbf{b}_{\perp_\text{2D}\mathbf{a}} = \begin{bmatrix}
 0 & 1 & 0 \\
 0 & 0 & 1
 \end{bmatrix}\mathbf{R}_{\mathbf{a}\rightarrow\mathbf{e}_x}\mathbf{b},
\end{equation}

We also define $\text{proj}_\mathbf{a}\mathbf{b}$ as the scalar projection of $\mathbf{b}$ onto $\mathbf{a}$. This component of $\mathbf{b}$ \emph{is not} affected by $\mathbf{a}$-axis rotation. Again, we can write this in matrix form as:
\begin{equation}
    \text{proj}_\mathbf{a}\mathbf{b} = \left[1,0,0\right]\mathbf{R}_{\mathbf{a}\rightarrow\mathbf{e}_x}\mathbf{b}.
\end{equation}

We can now use these operations to define the $SO(2)$-invariant inputs.

The directional part of the invariant input now contains three components: $\mathbf{d}_\mathbf{a}^\prime=(\text{proj}_\mathbf{a}\mathbf{d}, \langle \mathbf{d}_{\perp_\text{2D}\mathbf{a}},\mathbf{Z}_{\perp_\text{2D}\mathbf{a}} \rangle, \|\mathbf{d}_{\perp_\text{2D}\mathbf{a}}\|)$. The first is the invariant component of $\mathbf{d}$, i.e.~the part that is unaffected by an $\mathbf{a}$-axis rotation and can therefore be used directly. The second encodes $\mathbf{d}$ relative to $\mathbf{Z}$ in the plane perpendicular to $\mathbf{a}$. The third measures the norm of $\mathbf{d}$ projected into the plane perpendicular to $\mathbf{a}$, which is unchanged by rotations about $y$.

The conditioning part of the invariant input now contains two components: $\mathbf{Z}_\mathbf{a}^\prime=(\text{proj}_\mathbf{a}\mathbf{Z},\text{VN-In}(\mathbf{Z}_{\perp_\text{2D}\mathbf{a}}))$. The first is simply the invariant component of each column of $\mathbf{Z}$. The second is the Vector Neurons invariant layer transformation of the latent vectors projected into the plane perpendicular to $\mathbf{a}$.

\section{RENI++: A Statistical Model of Natural Illumination}\label{sec:RENI}

We now describe how to construct a statistical model of natural illumination environments as a rotation-equivariant conditional spherical neural field trained on a dataset of HDR outdoor illuminations. Natural environments have a canonical ''up'' direction (defined by gravity) but arbitrary rotation about this vertical axis. For this reason, in our model of natural illumination we do not want full $SO(3)$ rotation equivariance. This would have the undesirable effect of permitting unnatural environment orientations (such as with the sky at the bottom), providing a less useful prior when solving inverse problems. We therefore use the $SO(2)$ invariant formulation in Section \ref{sec:SO2} with the equivariant rotation axis set to the vertical ($y$) axis: $\mathbf{a}=\mathbf{e}_y$. We named our original model RENI (Rotation-Equivariant Natural Illumination) \cite{gardner2022rotation}. Here, we name our extended model RENI++ and explain below our new scale-invariant property, HDR representation, underlying neural field architecture and the training data and losses used in our implementation.

\subsection{Training Data} 
\label{Training Data}
HDR illumination is essential for realistic rendering and enables the accurate representation of the full dynamic range of natural light. Therefore our model must learn an HDR representation of natural illumination. We have curated a dataset of 1,694 HDR equirectangular images of outdoor, natural illumination environments obtained with a CC0 1.0 Universal Public Domain Dedication license \cite{poly_haven_hdris_nodate, giantcowfilms_hdris_nodate, ihdri_hdris_nodate, hdrmaps_hdris_nodate, whitemagus_3d_hdris_nodate, hdri_skies_hdris_nodate, textures_hdris_nodate}. All images were then checked to ensure they did not contain any personally identifiable information or offensive content and any images that contained predominantly unnatural light sources were removed. $21$ images were also selected and held back for optimising only the latent codes at test time, resulting in a training dataset of 1,673 HDR images. The data can then be augmented with horizontal reflection of the equirectangular images. Since a mirror image of an illumination environment is also a plausible illumination environment.

Each training batch comprises $P$ pairs of directions and corresponding log scaled $\text{log}(\text{HDR})$ RGB colours that we store in the matrices $\mathbf{D}=[\mathbf{d}_{1},\dots,\mathbf{d}_{P}]\in\R^{3\times P}$ and $\mathbf{C}=[\mathbf{c}_{1},\dots,\mathbf{c}_{P}]\in\R^{3\times P}$ respectively. Each batch contains samples distributed across all the images in the training set. As RENI++ is a continuous neural field, it is agnostic to the resolution and sampling of the spherical signal. The neural field can be queried for any direction. In practice, our dataset contains equirectangular spherical images. Because equirectangular images feature irregular sampling, we select directions according to the following criteria:
\begin{itemize}
    \item For the azimuthal angle \( \theta \), we sample uniformly in the interval \( [0, 2\pi] \):
\[
\theta \sim \mathcal{U}(0, 2\pi)
\]
\item For the polar angle \( \phi \), we sample according to a probability density function \( g(\phi) \) defined on \( [0, \pi] \):
\[
g(\phi) = \frac{\sin(\phi)}{2}
\]
Here, the division by 2 normalizes \( g(\phi) \) to ensure its integral is equal to 1.
\end{itemize}
In contrast to RENI \cite{gardner2022rotation}, in which each batch contained all rays from a single image and a sine-weighting in the loss function was used to compensate for the irregular sampling, we empirically found this new sampling procedure smoothed the loss landscape and removed the requirement for multi-resolution training. 

\subsection{High Dynamic Range Scale Invariance}
Computing a reconstruction loss in linear HDR space is dominated by large values and leads to a poor reconstruction of most of the environment. Therefore, similar to \cite{eilertsen_hdr_2017}, we train our network to output $\text{log}(\text{HDR})$ values and compute losses in log space. 

As the HDR images in our dataset are at unknown exposure values (EV) any image could be scaled by a global constant and would still represent a possible environment, simply one captured at a different EV. Therefore training RENI++ using an L2 reconstruction loss directly on pixel values, as was done in our prior work, results in over-fitting to the arbitrary EVs of images in the training data. This results in the latent space struggling to represent out-of-distribution EVs. To address this we take inspiration from monocular depth estimation techniques \cite{eigenDepthMapPrediction2014, liMegaDepthLearningSingleView2018} and train RENI++ using a scale-invariant loss:
\begin{equation}
  \mathcal{L}_{\text {scale-inv}}=\frac{1}{P} \sum_{p=1}^P\left(R_p\right)^2-\frac{1}{P^2}\left(\sum_{p=1}^P R_p\right)^2,
\end{equation}
where $R_{p} = f(\mathbf{d}_p^\prime, \mathbf{Z}^\prime_p) - \text{log}(\mathbf{c}_p)$ at pixel $p$ and $\mathbf{Z}^\prime_p$ is the invariant representation of the latent code of the image from which the $p$th pixel in the batch was drawn. This computes the mean square error (MSE) of the difference between all pairs of log-depths in linear time and results in the latent space of our model now representing a scale-free HDR image. Since the output of our model is only up to an unknown scale, we compute an optimal scale between reconstruction and ground truth before computing metrics. In practice, we do this using ordinary least squares in log space. We found that computing the optimal scale in linear space had no significant effect on the final metrics. In an inverse rendering setting, this scale becomes a per-image optimisable parameter representing the overall brightness of the environment.

To encourage accurate colour reproduction we also include a cosine similarity loss $\mathcal{L}_\text{cosine}$ on the RGB colour vectors:
\begin{equation}\label{cosine_similarity_loss}
    \mathcal{L}_{\text{cosine}} = 1.0 - \frac{1}{P} \sum_{i=1}^P \frac{f(\mathbf{d}_p^\prime, \mathbf{Z}^\prime_p) \cdot \text{log}(\mathbf{c}_p)}{\|f(\mathbf{d}_p^\prime, \mathbf{Z}^\prime_p)\| \|\text{log}(\mathbf{c}_p)\|}
\end{equation}

Since this only penalises the error in the RGB direction, it too is scale-invariant.

\subsection{Variational Auto-decoder}
We train our conditional spherical neural field as a decoder-only architecture, i.e. an auto-decoder \cite{park_deepsdf_2019} or Generative Latent Optimisation \cite{bojanowskiOptimizingLatentSpace2018}. This means that we optimise the network weights simultaneously with the latent codes for each training sample. This avoids the need to design a rotation-equivariant encoder while, for inverse tasks, only the decoder is needed so we avoid the redundancy of also training an encoder. However, training with no regularisation on the learnt latent space does not lead to a space that is smooth or that follows a known distribution. This means it cannot be sampled from, does not produce meaningful interpolations and provides no prior for inverse problems. For this reason, we use a \emph{variational auto-decoder} \cite{zadeh_variational_2021} architecture. Each training sample is represented by a mean, $\boldsymbol{\mu}_{i}\in\R^{3N}$, and standard deviation, $\boldsymbol{\sigma}_{i}\in\R^{3N}$, that provide the parameters of a normal distribution from which the flattened latent code for that training sample is drawn: $\text{vec}(\mathbf{Z}_i)\sim\mathcal{N}(\boldsymbol{\mu}_{i},\boldsymbol{\Sigma}_i)$, where $\boldsymbol{\Sigma}_i=\text{diag}(\sigma_{i,1}^2,\dots,\sigma_{i,3N}^2)$ is the diagonal covariance matrix. Using the reparameterisation trick \cite{kingma_auto-encoding_2013}, we can generate a latent code \(\text{vec}(\mathbf{Z}_{i}) = \boldsymbol{\mu}_{i}+ \boldsymbol{\sigma}_{i} \odot \boldsymbol{\epsilon}\) where the noise is sampled as \(\boldsymbol{\epsilon} \sim \mathcal{N}(\mathbf{0}, \mathbf{I}_{3N})\). During training, we optimise $\boldsymbol{\mu}_{i}$ and  $\boldsymbol{\sigma}_{i}$ for each training sample and use the Kullback–Leibler divergence (KLD) as a loss to regularise the distribution of each latent code toward the standard normal distribution:
\begin{equation}\label{kld_loss}
    \mathcal{L}_{\text{KLD}}= -\frac{1}{2}\sum_{i=1}^{K}\frac{1}{D}\sum_{j=1}^{D}(1 + \log(\sigma_{i,j}^{2})-\mu_{i,j}^{2} - \sigma_{i,j}^{2}),
\end{equation}
where $K$ is the number of unique latent codes in a batch, and $D = 3N$ is the dimensionality of the latent space.

\subsection{Neural Field} 
Our neural field decoder, $f$, is conditioned using cross-attention layers as per \cite{rebainAttentionBeatsConcatenation2023}. Similar to the Transformer decoder \cite{vaswaniAttentionAllYou2017} architecture, our neural field operates as follows.

For each layer \( l \) in the decoder, we apply multi-head attention (MHA):
\begin{equation}
    \text{MHA}^l(\mathbf{Q}, \mathbf{K}, \mathbf{V}) = \text{Concat}(\text{head}_1^l, \ldots, \text{head}_h^l)W^O + \mathbf{Q},
\end{equation}
and a position-wise feed-forward network (FFN):
\begin{multline}
\text{FFN}^l(\mathbf{d}'^{(l-1)}) = \text{LN}(\mathbf{d}'^{(l-1)}) \\
\quad + \text{ReLU}(\text{LN}(\mathbf{d}'^{(l-1)})W_1^l + b_1^l)W_2^l + b_2^l,
\end{multline}
where $ \mathbf{Q} = \mathbf{d}^\prime W^Q $, $ \mathbf{K} = \mathbf{Z}^\prime W^K $, and $ \mathbf{V} = \mathbf{Z}^\prime W^V $ and $\text{LN}$ is the layer norm operation. Here, $ \mathbf{d}^\prime $ is the output from the previous layer or the original input for the first layer, $ \mathbf{Z}^\prime $ is the conditioning input, and the heads correspond to different projections of these inputs. Residual connections are applied around both the MHA and FFN modules, with layer normalisation following each residual connection.

The output of the final layer $L$ is then passed through an output projection to predict HDR RGB values:
\begin{equation}
\text{Output} = \sigma(\text{FFN}^L(\mathbf{d}'^L)W^{\text{out}} + b^{\text{out}}).
\end{equation}
An optional non-linearity $ \sigma $ activation function is applied at the output stage.

The original RENI model \cite{gardner2022rotation} used \emph{conditioning-by-concatenation} \cite{sitzmann_implicit_2020, xie_neural_2021}, i.e.~both the directional, $\mathbf{d}^\prime$, and conditioning, $\mathbf{Z}^\prime$, inputs were passed directly as input to the first layer of the decoder network. This decoder network was a SIREN \cite{sitzmann_implicit_2020}, an MLP with a periodic activation function, as this architecture had proven highly effective at representing a wide range of complex natural signals. However similar to \cite{rebainAttentionBeatsConcatenation2023} we found attention with positional encoding on directions outperformed the \emph{conditioning-by-concatenation} SIREN and \cite{sitzmann_implicit_2020, xie_neural_2021} and \emph{FiLM-conditioned} \cite{chan_pi-gan_2021} SIREN when fitting latent codes to unseen images.

\begin{figure}
    \centering
    \makebox[\columnwidth]{%
        \begin{tikzpicture}
        % \draw[step=0.5cm,gray,very thin] (-3,-2) grid (3,2);
        \node (img) {\includegraphics[width=0.90\columnwidth]{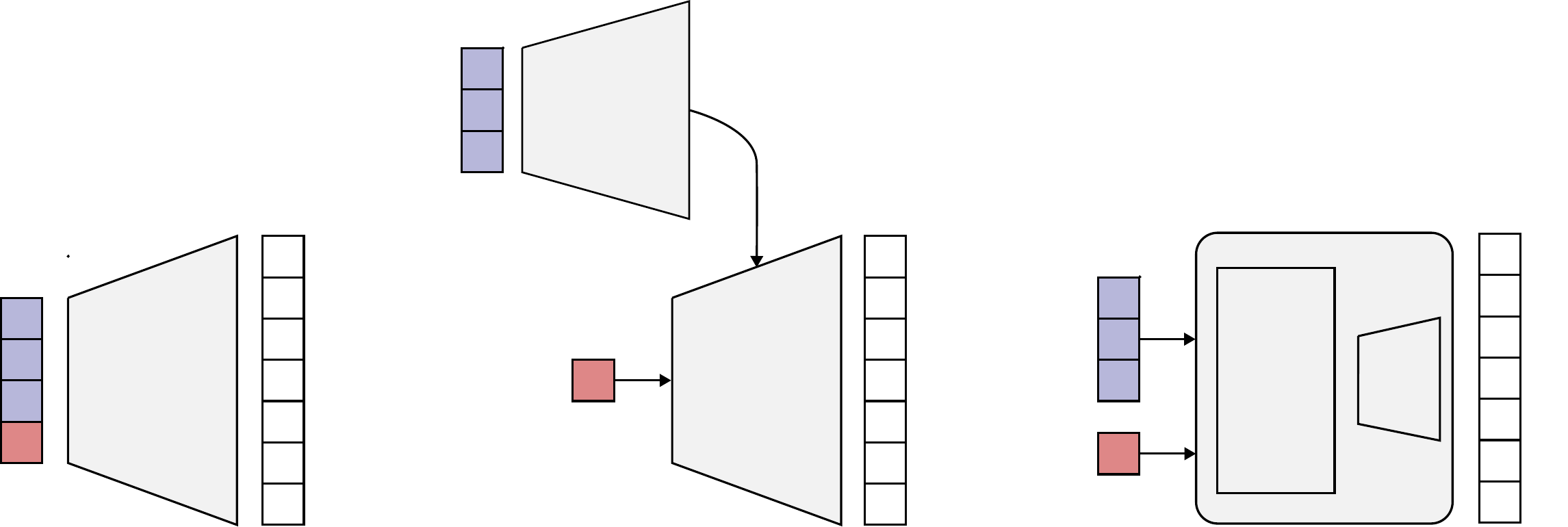}};
    
        % Add 'outputs' labels rotated by 90 degrees
        \node[rotate=90, font=\tiny] at (-2.29, -0.60) {outputs};
        \node[rotate=90, font=\tiny] at (0.76, -0.60) {outputs};
        \node[rotate=90, font=\tiny] at (3.9, -0.60) {outputs};
    
        % Add 'z' labels for the orange inputs
        \node[font=\tiny] at (-4.15, -0.48) {z};
        \node[font=\tiny] at (-1.79, 0.8) {z};
        \node[font=\tiny] at (1.45, -0.38) {z};
    
        % Add 'λ(x)' labels for the red inputs
        \node[font=\tiny] at (-4.3, -0.92) {$\lambda(x)$};
        \node[font=\tiny] at (-1.38, -0.59) {$\lambda(x)$};
        \node[font=\tiny] at (1.3, -0.97) {$\lambda(x)$};
    
        % Add 'MLP' in each of the networks
        \node[font=\tiny] at (-3.2, -0.59) {MLP};
        \node[font=\tiny] at (-0.1, -0.59) {MLP};
        \node[rotate=90, font=\tiny] at (3.15, -0.59) {MLP};
    
        % Add 'MLP' in each of the networks
        \node[font=\tiny] at (-3.2, -1.7) {Condition-by-Concat};
        \node[font=\tiny] at (0.0, -1.7) {Hypernetwork};
        \node[font=\tiny] at (2.8, -1.7) {Attention};
    
        % Extra
        \node[font=\tiny] at (-0.85, 0.95) {Hyper};
        \node[font=\tiny] at (-0.85, 0.65) {MLP};
        \node[font=\tiny] at (-0.0, 0.75) {\textbf{W}};
        \node[font=\tiny] at (2.35, -0.24) {K};
        \node[font=\tiny] at (2.35, -0.50) {V};
        \node[font=\tiny] at (2.35, -0.97) {Q};
    
        \end{tikzpicture}
    }
    \caption{Three methods of conditioning in Neural Fields. Similar to \cite{rebainAttentionBeatsConcatenation2023} we found \emph{attention} outperformed \emph{conditioning-by-concatenation} and \emph{FiLM-conditioning} in capturing high-frequency details.}
    \label{fig:model_types}  
\end{figure}

\subsection{Training} \label{Training Details}
RENI++ is implemented in Nerfstudio \cite{nerfstudio}, a framework for developing Neural Field based applications. We use Adam \cite{kingma_adam_2015}, along with a Cosine Decay \cite{loshchilov2017sgdr} schedule and 500-step 'warm-up' phase, to optimise the sum of the scale-invariant and KLD losses:
\begin{equation}\label{training_loss}
    \mathcal{L}_{\text{Train}} = \rho \mathcal{L}_{\text{scale-inv}} + \gamma \mathcal{L}_{\text{cosine}} + \beta \mathcal{L}_{\text{KLD}}
\end{equation}
We randomly initialise the mean latent code for each image, \(\boldsymbol{\mu}_{i}\), from a standard normal distribution. In order to ensure the positivity of the variances, $\sigma^2_{i,j}$, we optimise $\log(\sigma^2_{i,j})$ which we initialise randomly from a normal distribution with mean -5 and variance 1. With the new sampling procedure and model architecture described above we found better performance than the conference paper even when training for $80x$ fewer steps. We found the best performance when using a transformer-based variational auto-decoder with $8$ attention heads $6$ decoder layers and $128$ parameters per layer. We use a cosine decaying learning rate \cite{loshchilov2017sgdr} with a warm-up phase modelled as:

$$\text{lr} = \text{lr}_{0} \times \frac{s}{\text{W}}$$

Where $W$ is the number of warm-up steps, $\text{lr}_{0}$ is the initial learning rate and $s$ is the current step. The post-warm-up learning rate is then computed via:

$$\text{lr} = \text{lr}_{0} \times \left( \alpha + \frac{1 - \alpha}{2} \times (\cos(\pi \times \frac{s - \text{W}}{s_{\text{max}} - \text{W}}) + 1) \right) $$

Where $s_\text{max}$ is the maximum number of training steps and $\alpha$ is the minimum learning rate as a fraction of the initial learning rate. We use a $500$-step warm-up phase, $\alpha = 5 \times 10^{-2}$ and an initial learning rate of \(10^{-3}\). Our loss weights are \(\rho=1.0\), \(\gamma=1.0\) and \(\beta=10^{-6}\). Training RENI++ with a latent code dimension of $D = 27$, $\text{num heads} = 8$, $\text{num layers} = 6$, $\text{hidden features} = 128$ and NeRF \cite{mildenhall_nerf_2020} positional encoding on the directions takes around 15 minutes on an Nvidia RTX 3080-Ti 16GB Laptop GPU.

\subsection{Model Fitting} At test time the decoder is held static and only latent codes are optimised to fit an unseen image. We initialise the latent code to zeros, corresponding to the mean environment map (see Figure \ref{fig:teaser}, left). This provides an unbiased initialisation in the absence of any prior information about the environment. We use the same losses as used during training except we remove $\mathcal{L}_{\text{KLD}}$. Our test time loss is therefore:
\begin{equation}\label{test_loss}
    \mathcal{L}_{\text{Test}} = \rho \mathcal{L}_{\text{scale-inv}} + \gamma \mathcal{L}_{\text{Cosine}}
\end{equation}
We found the best performance when using $\rho = 1.0$, $\gamma = 1.0$, and Adam with an exponentially decaying learning rate starting at \(10^{-1}\) and decreasing to \(10^{-7}\) over $2500$ steps. This takes around $1$ minute to fit all $21$ images in the test dataset on an Nvidia RTX 3080-Ti 16GB Laptop GPU.

A PyTorch implementation, our dataset and trained models can be found at \href{https://github.com/JADGardner/ns_reni}{github.com/JADGardner/ns\_reni}.

\section{Evaluation}\label{sec:eval}

\subsection{Generalisation}
\label{Generalisation} 

\begin{table*}\centering
\ra{1.3}
\begin{tabular}{@{}lrrrrrrrrrrrrrrr@{}}\toprule
& \multicolumn{3}{c}{RENI} & \phantom{ab}& \multicolumn{3}{c}{RENI++} &
\phantom{ab} & \multicolumn{3}{c}{SH} &
\phantom{ab} & \multicolumn{3}{c}{SG}\\
\cmidrule{2-4} \cmidrule{6-8} \cmidrule{10-12} \cmidrule{14-16}
D & PSNR$\uparrow$ & SSIM$\uparrow$ & LPIPS$\downarrow$ && PSNR$\uparrow$ & SSIM$\uparrow$ & LPIPS$\downarrow$ && PSNR$\uparrow$ & SSIM$\uparrow$ & LPIPS$\downarrow$ && PSNR$\uparrow$ & SSIM$\uparrow$ & LPIPS$\downarrow$\\ \midrule
27 & 
17.02 & \textbf{0.40} & 0.75 &&
\textbf{18.02} & 0.39 & \textbf{0.62} &&
15.00 & 0.33 & 0.79 && 
17.67 & 0.35 & 0.80 \\
108 & 
19.58 & 0.45 & 0.75 && 
\textbf{20.87} & \textbf{0.49} & \textbf{0.56} &&
17.88 & 0.38 & 0.79 && 
19.27 & 0.41 & 0.70 \\
147 & 
19.97 & 0.46 & 0.74 && 
\textbf{21.13} & \textbf{0.51} & \textbf{0.55} && 
18.23 & 0.40 & 0.77 && 
19.81 & 0.43 & 0.67 \\ 
300 & 
20.47 & 0.48 & 0.73 && 
\textbf{22.10} & \textbf{0.55} & \textbf{0.52} && 
19.15 & 0.45 & 0.74 && 
20.24 & 0.46 & 0.61 \\
\bottomrule
\end{tabular}
\caption{The mean PSNR, SSIM, and LPIPS scores for tone-mapped LDR images when fitting to the test set for increasing latent dimensions. Where an exact dimensionality comparison is not possible for SG, e.g. in the $D=27$ and $D=147$ cases, we bias in favour of SG with a $D=30$ and $D=150$ respectively.}
\label{tab:comparison_PSNR_SSIM_LPIPS}
\end{table*}

\begin{figure*}
    \centering
    \makebox[\textwidth]{%
        \begin{tikzpicture}
            \node (img) {\includegraphics[width=1.00\textwidth]{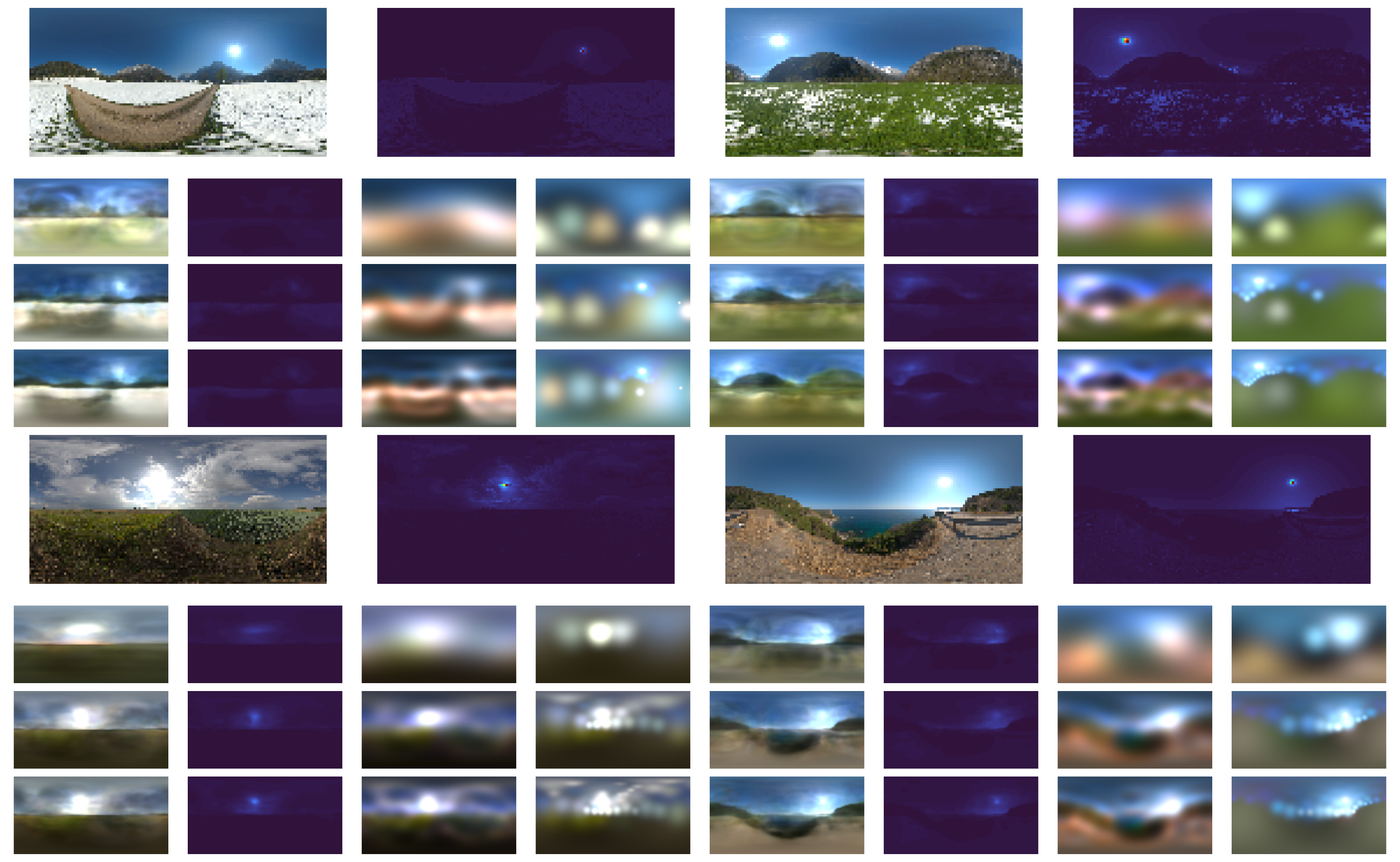}};
            \node at (-9.1, 4.5) [rotate=90] {\tiny Ground Truth};
            \node at (-9.1, -1.0) [rotate=90] {\tiny Ground Truth};
            \node at (-9.1, 2.75) [rotate=90] {\tiny 27};
            \node at (-9.1, 1.65) [rotate=90] {\tiny 147};
            \node at (-9.1, 0.55) [rotate=90] {\tiny 300};
            \node at (-9.1, -2.8) [rotate=90] {\tiny 27};
            \node at (-9.1, -3.9) [rotate=90] {\tiny 147};
            \node at (-9.1, -5.0) [rotate=90] {\tiny 300};
            \node at (-6.8, 3.4) {\tiny RENI++};
            \node at (-3.4, 3.4) {\tiny SH};
            \node at (-1.1, 3.4) {\tiny SG};
            \node at (2.3, 3.4) {\tiny RENI++};
            \node at (5.65, 3.4) {\tiny SH};
            \node at (7.9, 3.4) {\tiny SG};
            \node at (-6.8, -2.15) {\tiny RENI++};
            \node at (-3.4, -2.15) {\tiny SH};
            \node at (-1.1, -2.15) {\tiny SG};
            \node at (2.3, -2.15) {\tiny RENI++};
            \node at (5.65, -2.15) {\tiny SH};
            \node at (7.9, -2.15) {\tiny SG};
        \end{tikzpicture}
    }
    \caption{Generalisation to unseen images with latent code dimensions, $D = 3N$ for $N=9, 36, 49$ and for SH of equal dimensionality (orders 2, 6, and 9). SG results are with dimensionality $D = 30, 150, 300$. Log-scale heat maps for ground truth and RENI++ are also shown.}
    \label{fig:comparison}
\end{figure*}

We begin by evaluating the generalisation ability of RENI++ when approximating unseen environments. We compare against original RENI, SH and SG and explore how generalisation performance varies as a function of latent code dimension. We consider RGB spherical harmonics of order 2, 5, 6 and 9 equal to latent code dimension $D = 3 \times N$ for $N=9, 36, 49, 100$. Since SG requires a dimensionality that is a multiple of 6, where $D$ is not a multiple of 6, we bias in SGs favour by using the next multiple, i.e.~$\lceil D/6 \rceil \cdot 6$. We re-implemented within Nerfstudio open-source implementations of per-pixel environment map fitting for SG provided by \cite{li_inverse_2020} and \cite{charlmersgitSphericalHarmonics2022} for SH and used these to fit our models within the same environment as RENI++. As shown in Figure \ref{fig:comparison}, RENI++ can capture higher frequency detail than SH, SG and the original RENI implementation, is less dominated by high-value pixels and can reproduce accurate HDR values. A comparison of the mean PSNR across the test set images for increasing latent code dimensionality is shown in Table \ref{tab:comparison_PSNR_SSIM_LPIPS}.

\subsection{Latent Space Interpolation}
\begin{figure*}[!ht]
    \centering
    \begin{tikzpicture}
    \node (img) {\includegraphics[width=0.98\textwidth]{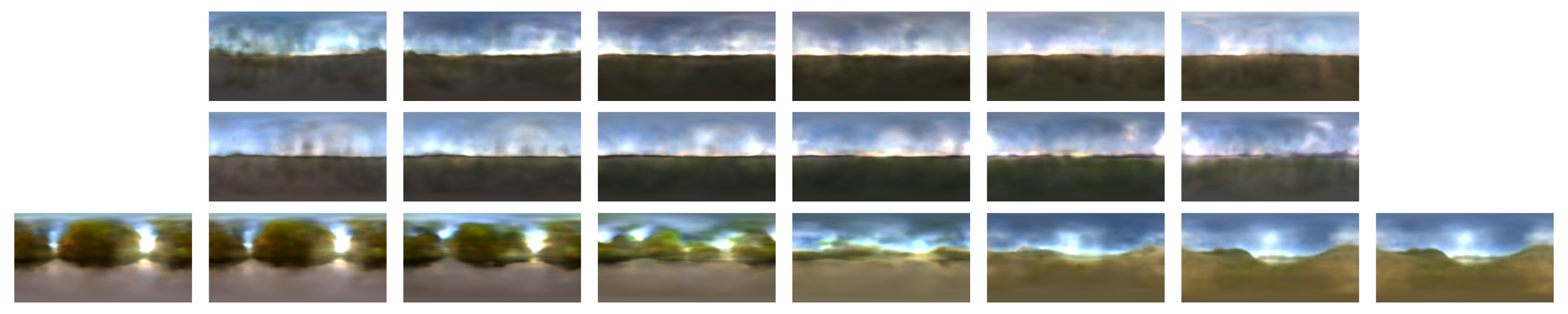}};
    \end{tikzpicture}
    \caption{Interpolation results for RENI++ with latent code dimension of $D = 300$. Rows 1 and 2 show interpolations between two random latent codes, and row 3 shows an interpolation between two training images with the ground-truth images shown.}
    \label{fig:Interpolations}
\end{figure*}

As shown in Figure \ref{fig:Interpolations}, linear interpolations between codes result in smooth transitions between images and plausible natural environments for all intermediate latent codes showing how RENI++ encodes a meaningful internal representation of natural illumination.

\subsection{Environment Completion}

\begin{figure*}[h]
    \centering
    \begin{tikzpicture}
    \node (img) {\includegraphics[width=0.98\textwidth]{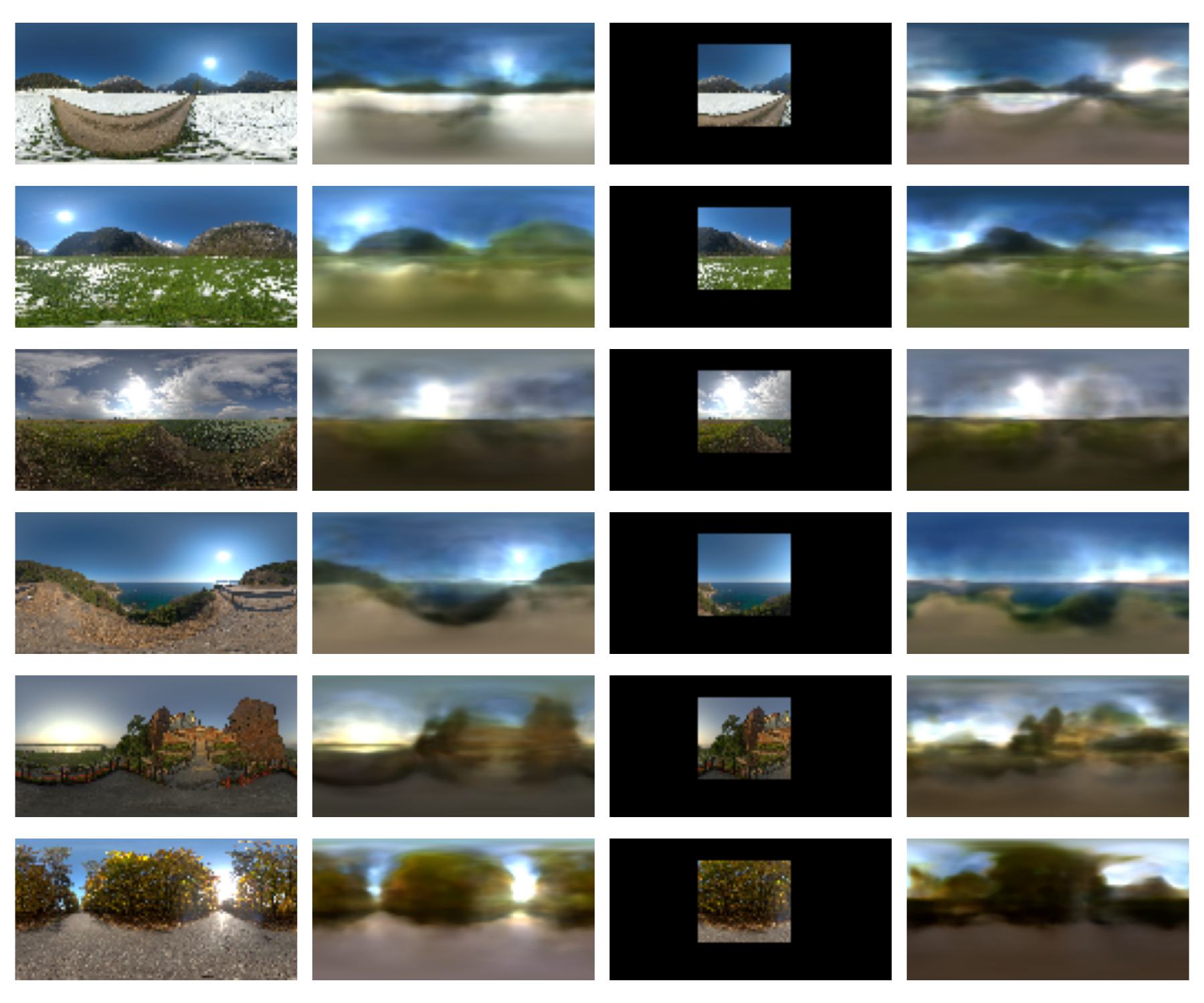}};
    \node at (-6.7, 7.65) {\small Ground};
    \node at (-6.7, 7.30) {\small Truth};
    \node at (-2.25, 7.65) {\small RENI++};
    \node at (-2.25, 7.30) {\small Full};
    \node at (2.2, 7.65) {\small Masked};
    \node at (2.2, 7.30) {\small Ground Truth};
    \node at (6.6, 7.65) {\small RENI++};
    \node at (6.6, 7.30) {\small Outpainting};
    \end{tikzpicture}
    \caption{Col. $2$ shows the output of optimising latent codes on full ground-truth (Col. $1$). When trained on a masked ground-truth (Col. $3$) RENI++ predicts plausible continuations for the environment and makes accurate estimations of sun locations (Col. $4$). Results from a \(D = 300\) model.}
    \label{fig:outpainting}
\end{figure*}

Any picture of a natural scene contains cues about the surrounding environment that the scene was captured in, such as likely sun locations and possible environmental content. If RENI++ is only provided with a small portion of the complete environment map in its loss at test time, RENI++ can hallucinate plausible completions of the environment. As shown in Figure \ref{fig:outpainting}, RENI++ makes sensible estimations about the possible colours and shapes of land and sky and often predicts quite accurate sun locations despite the sun being outside the image crop.

\begin{figure*}[h]
    \centering
    \begin{tikzpicture}
    \node (img) {\includegraphics[width=0.98\textwidth]{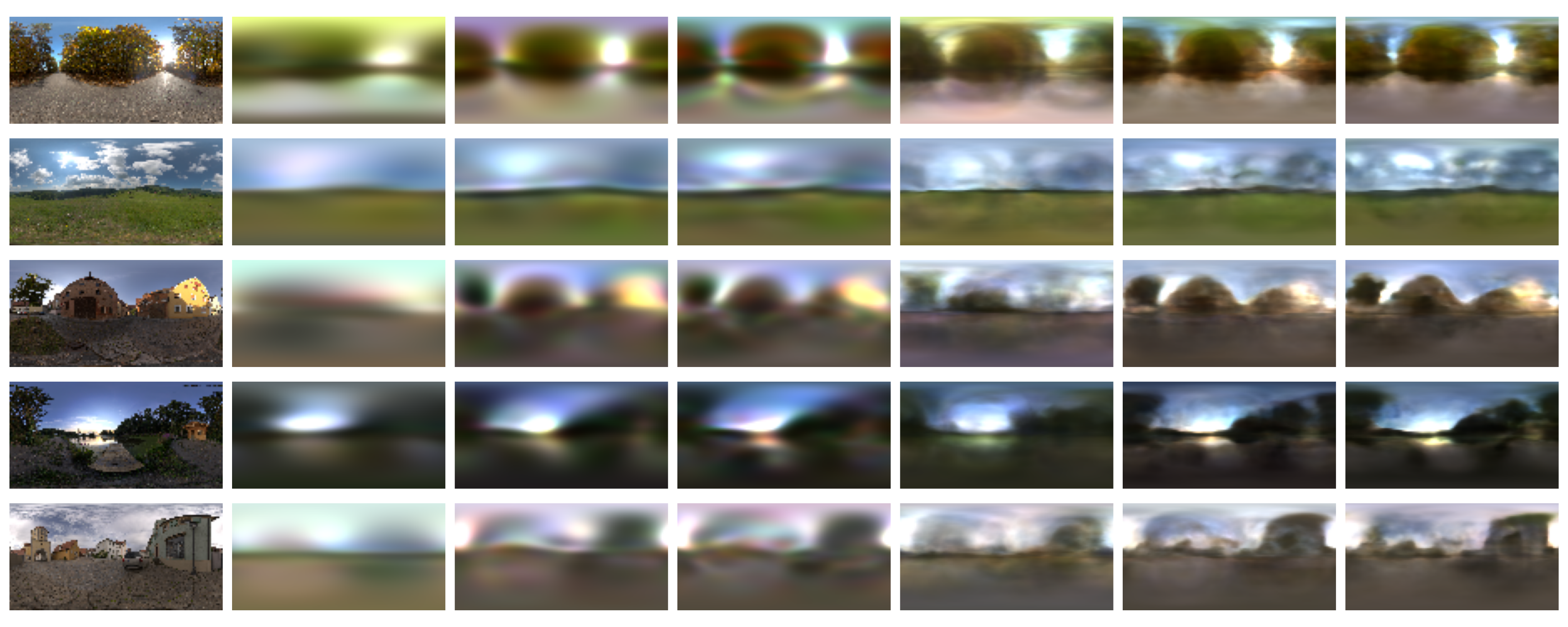}};
    \node at (-7.65, 3.95) {\small Ground};
    \node at (-7.65, 3.60) {\small Truth};
    \node at (-2.55, 3.95) {\small RENI};
    \node at (5.1, 3.95) {\small RENI++};
    \node at (-5.1, 3.60) {\small 27};
    \node at (-2.55, 3.60) {\small 147};
    \node at (0.0, 3.60) {\small 300};
    \node at (2.5, 3.60) {\small 27};
    \node at (5.0, 3.60) {\small 147};
    \node at (7.55, 3.60) {\small 300};
    \end{tikzpicture}
    \caption{A comparison between our prior RENI implementation and RENI++ for various latent dimensions. Col. $1$ shows the ground-truth environment map. Cols. $2-7$ shows prior RENI and RENI++ output for model sizes \(D = 27, 149, 300\).}
    \label{fig:old_vs_new}
\end{figure*}

\begin{figure*}[!hb]
  \centering
  \makebox[\textwidth]{%
      \begin{tikzpicture}
        \newcommand\smallermath{\fontsize{4pt}{5pt}\selectfont}
        \node (img) {\includegraphics[width=\textwidth]{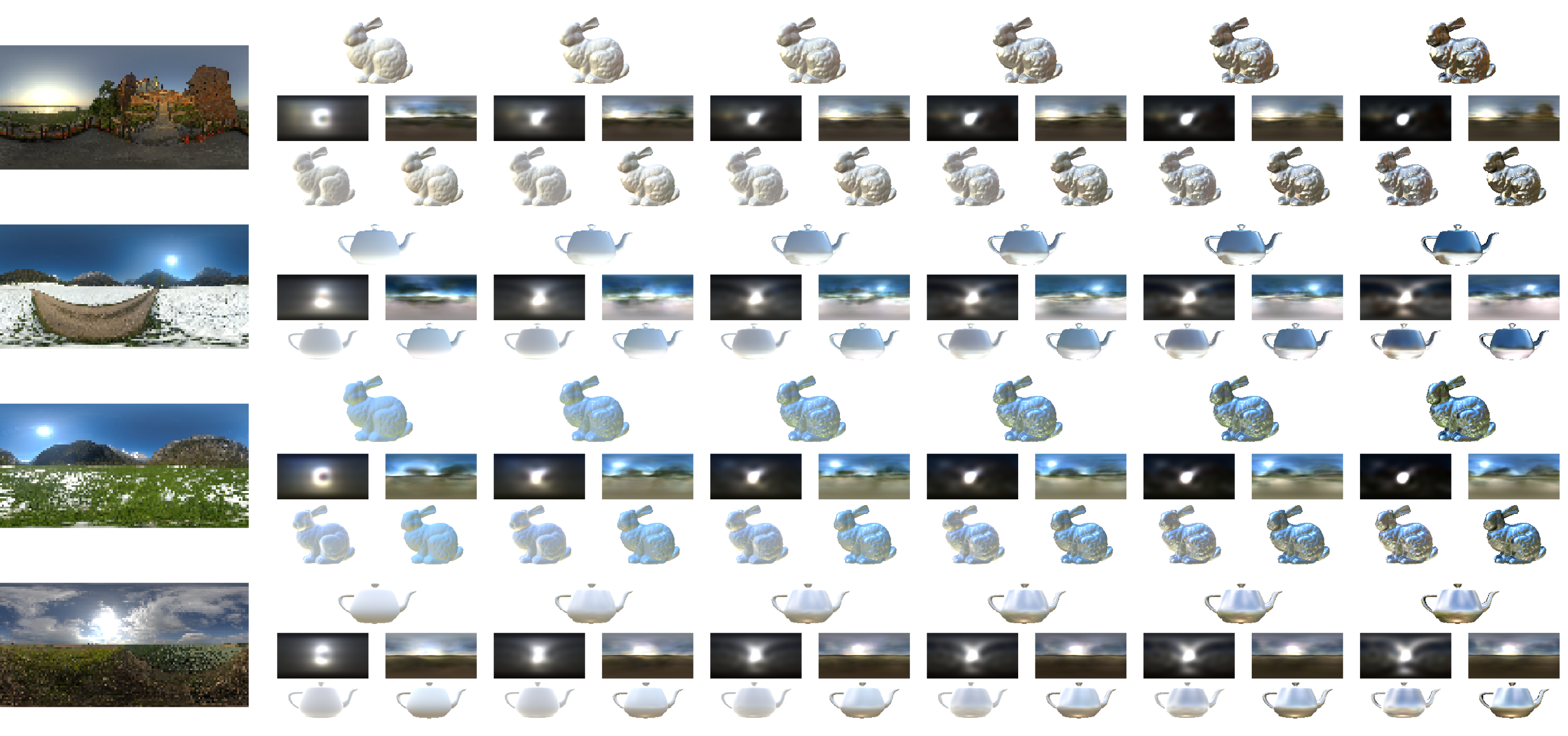}};
        \node at (-7.65, 4.1) {\tiny Ground Truth};
        \node at (-7.65, 3.85) {\tiny Environment Map};
        \node[rotate=90, font=\smallermath] at (-6.05, 3.6) {Target};
        \node[rotate=90, font=\smallermath] at (-6.05, 2.9) {Recon};
        \node[rotate=90, font=\smallermath] at (-6.05, 2.2) {Fit};
        
        \node[rotate=90, font=\smallermath] at (-6.05, 1.45) {Target};
        \node[rotate=90, font=\smallermath] at (-6.05, 0.75) {Recon};
        \node[rotate=90, font=\smallermath] at (-6.05, 0.05) {Fit};
        
        \node[rotate=90, font=\smallermath] at (-6.05, -0.6) {Target};
        \node[rotate=90, font=\smallermath] at (-6.05, -1.3) {Recon};
        \node[rotate=90, font=\smallermath] at (-6.05, -2.0) {Fit};
        
        \node[rotate=90, font=\smallermath] at (-6.05, -2.65) {Target};
        \node[rotate=90, font=\smallermath] at (-6.05, -3.35) {Recon};
        \node[rotate=90, font=\smallermath] at (-6.05, -4.05) {Fit};
    
        \node[font=\smallermath] at (1.5, 4.55) {Increasing Specularity};
        
        \draw[->] (-5.8,4.4) -- (8.8,4.4); % Draws a line with an arrow at the end
    
        \node[font=\smallermath] at (-4.75, 4.2) {0.0};
        \node[font=\smallermath] at (-2.25, 4.2) {0.2};
        \node[font=\smallermath] at (0.3, 4.2) {0.4};
        \node[font=\smallermath] at (2.8, 4.2) {0.6};
        \node[font=\smallermath] at (5.3, 4.2) {0.8};
        \node[font=\smallermath] at (7.8, 4.2) {1.0};
    
        % For 0.0
        \node[font=\smallermath] at (-5.3, 3.25) {SH};
        \node[font=\smallermath] at (-4.1, 3.25) {RENI++};
        
        % For 0.2
        \node[font=\smallermath] at (-2.8, 3.25) {SH};
        \node[font=\smallermath] at (-1.6, 3.25) {RENI++};
        
        % For 0.4
        \node[font=\smallermath] at (-0.3, 3.25) {SH};
        \node[font=\smallermath] at (0.9, 3.25) {RENI++};
        
        % For 0.6
        \node[font=\smallermath] at (2.2, 3.25) {SH};
        \node[font=\smallermath] at (3.4, 3.25) {RENI++};
        
        % For 0.8
        \node[font=\smallermath] at (4.7, 3.25) {SH};
        \node[font=\smallermath] at (5.9, 3.25) {RENI++};
        
        % For 1.0
        \node[font=\smallermath] at (7.2, 3.25) {SH};
        \node[font=\smallermath] at (8.4, 3.25) {RENI++};

      \end{tikzpicture}
  }
  \caption{Reconstruction results in an inverse rendering task. The specular Blinn-Phong term $K_{s}$ increases from left to right in steps of $0.2$. Both RENI++ and SH have a dimensionality of $D = 300$. RENI++ outperforms SH across all $K_{s}$ and as $K_{s}$ increases the environments predicted by RENI++ become significantly more detailed and accurate.}
  \label{fig:inverse_rendering}
\end{figure*}

\begin{table*}[h]
\centering
\ra{1.3}
\begin{tabular}{@{}cccc@{}}\toprule
\textbf{Rotation Angle (Degrees)} & \textbf{Relative Error} & \textbf{Ground Truth Rotation Matrix Error} \\ \midrule
5 & $0.582 \pm 0.140$ & $0.150 \pm 0.108$ \\
20 & $0.762 \pm 0.140$ & $0.243 \pm 0.123$ \\
45 & $0.838 \pm 0.098$ & $0.416 \pm 0.291$ \\
90 & $0.049 \pm 0.049$ & $0.008 \pm 0.009$ \\
180 & $0.050 \pm 0.045$ & $0.007 \pm 0.005$ \\
270 & $0.044 \pm 0.048$ & $0.006 \pm 0.005$ \\
\bottomrule
\end{tabular}
\caption{Mean relative error and rotation matrix discrepancy for optimised latent codes across various rotation angles of the test set.}
\label{tab:rotation_error_comparison}
\end{table*}

\begin{table*}\centering
\ra{1.3}
\begin{tabular}{@{}c c cc cc cc cc@{}}\toprule
& & \multicolumn{2}{c}{27} & \multicolumn{2}{c}{108} & \multicolumn{2}{c}{147} & \multicolumn{2}{c}{300}   \\
\cmidrule(lr){3-4} \cmidrule(lr){5-6} \cmidrule(lr){7-8} \cmidrule(lr){9-10}
Component & Training Steps & LDR & HDR & LDR & HDR & LDR & HDR & LDR & HDR \\
\midrule
RENI \cite{gardner2022rotation} & 4,015,200 & 17.02 & 32.70 & 19.58 & 33.63 & 19.97 & 33.78 & 20.47 & 34.03 \\
RENI \cite{gardner2022rotation} & 50,000 & 12.93 & 30.94 & 14.18 & 31.41 & 15.54 & 31.90 & 16.53 & 32.00 \\
+ Data Augmentation & 50,000 & 12.18 & 30.87 & 14.40 & 31.44 & 15.46 & 31.82 & 15.80 & 31.91 \\
w/ Transformer Decoder & 50,000 & 13.05 & 30.90 & 14.71 & 31.52 & 15.06 & 31.81 & 16.90 & 32.28 \\
+ Scale-Invariance (RENI++) & 50,000 & \textbf{18.02} & \textbf{33.00} & \textbf{20.87} & \textbf{34.30} & \textbf{21.13} & \textbf{34.60} & \textbf{22.10} & \textbf{35.10} \\
\bottomrule
\end{tabular}
\caption{Mean PSNR in Low Dynamic Range (LDR) and High Dynamic Range (HDR) on the test set for various ablations of the RENI++ architecture. RENI is the model architecture and weights directly from the conference paper, converted to work in Nerfstudio, this was then training again to match the number of steps required to fit RENI++. The transformer decoder is sized to have a similar number of parameters as the SIREN-based condition-by-concatenation decoder. All results for models training within Nerfstudio are trained for $80x$ fewer steps than our prior RENI implementation. Despite this disadvantage, once Scale-Invariance is introduced, the Transformer Decoder outperforms RENI across all latent code sizes.}
\label{tab:ablation_study}
\end{table*}

\subsection{Inverse Rendering}
\label{Inverse Rendering} 

To test the performance of RENI++ in an inverse rendering pipeline, we implemented a normalised Blinn-Phong environment map shader within Nerfstudio, enabling fully differentiable rendering. We did not evaluate our model on the OpenIllumination dataset \cite{isabellaliulinghaochenOpenIlluminationMultiIlluminationDataset2023} due to its use of non-natural illumination conditions that RENI++ is not trained to represent. Instead, we render a 3D object with fixed geometry, pose, camera and material parameters such that only the lighting in the scene is unknown. Optimising only latent codes, we minimise Mean Squared Error and Cosine Similarity losses between a rendering using a ground truth environment map and one using the output of RENI++ or Spherical Harmonics. In RENI++'s case we also include a prior loss on the latent codes:
\begin{equation}\label{inverse_loss}
    \mathcal{L}_{\text{Inverse}} = \rho \mathcal{L}_{\text{MSE}} + \gamma \mathcal{L}_{\text{Cosine}} + \beta \mathcal{L}_{\text{Prior}} 
\end{equation}
where:
\begin{equation}\label{prior_loss}
    \mathcal{L}_{\text{prior}} = \frac{1}{K} \sum_{i=1}^K \left( \mathbf{Z}_i \right)^2
\end{equation}

with $K$ being the number of unique latent codes in a batch.
This was tested for incremental increases in the weighting of the Blinn-Phong specular term $(K_{s})$, from $K_{s} = 0$ to $K_{s} = 1.0$ in steps of $0.2$. A normalisation factor \(\zeta = (n+2) / (4\pi(2-e(\frac{-n}{2})))\) \cite{gotanda_physically-based_2010}, was applied to $K_{s}$ to get a steady transition from diffuse to specular. We achieved the best performance using $\rho = 10^{2}$, $\gamma = 1.0$ and $\beta = 10^{-3}$, and a learning rate of \(10^{-2}\) for 200 steps. We use an environment map resolution of $H=64$ and render the object with a resolution of $128^{2}$. In Figure \ref{fig:inverse_rendering} we compare against SH environment maps implemented within Nerfstudio and optimise the SH parameters using the same inverse rendering pipeline as the RENI++ latent codes. RENI++ outperforms SH for all $K_{s}$ values whilst, due to RENI++'s prior, ensuring realistic and natural-looking environment estimations.

\subsection{Non-convexity of Reconstruction Error in Latent Space}
\label{Non-convexity} 
RENI++ is rotation equivariant. However, this does not necessarily mean that optimising to fit a rotated image will yield a rotated version of the latent code resulting from fitting to an unrotated version of the image. This means that the loss landscape of our reconstruction losses is not convex. To verify this, we fit RENI++ to versions of the test set at different rotations. Initialising at the mean environment and optimising resulted in sets of latent codes $\mathbf{Z}_\text{Rot}$ that we could compare to latent codes fit to the unrotated images  $\mathbf{Z}_\text{Un-rot}$. Ideally, this would result in each latent code being explained via a $y$-axis rotation of the other. To test this we minimised $\mathbf{M}$ for $\left \| \mathbf{Z}_\text{Un-rot}\mathbf{M} - \mathbf{Z}_\text{Rot}  \right \|_{2}$ and obtained a rotation matrix $\mathbf{R}$ that minimises $\left \| \mathbf{M} - \mathbf{R} \right \|_{F}$. We then calculate the relative error between $\mathbf{R}\mathbf{Z}_\text{Un-rot}$ and $\mathbf{Z}_\text{Rot}$: 
\[
E = \frac{\left \| \mathbf{R}\mathbf{Z}_\text{Un-rot} - \mathbf{Z}_\text{Rot} \right \|_{F}}{\left \| \mathbf{Z}_\text{Rot} \right \|_{F}}
\]

The results are shown in Table. \ref{tab:rotation_error_comparison}. For the rotations of $90$, $180$, and $270$ degrees, the relative error is very small demonstrating that both latent codes can largely be explained as a simple rotation of the other. However, for the remaining three smaller rotations the error was higher, suggesting there is redundancy in the latent space, i.e. there are multiple possible explanations for a single image. Better latent space regularisation, tuning model dimensionality and a larger dataset might help resolve this.

\subsection{Equivariance Ablation}
\label{AblationEquivariance}
To test the impact of restricting RENI++'s equivariance to $SO(2)$ we ran an ablation of models with $SO(3)$, $SO(2)$ and without equivariance at three sizes of latent code dimension $D$. For the model without equivariance, we augmented the dataset with rotations of the images at increments of $0.785 rad$ for a training dataset size of $13384$ images. The $SO(2)$ case performs best for all latent code sizes, and both the $SO(2)$ and $SO(3)$ outperform the model trained purely using augmentation whilst using significantly less data. Results are shown in Table \ref{tab:comparison_equivariance}.

\begin{table}\centering
\begin{tabular}{@{}c ccc@{}}\toprule
D & None & SO(2) & SO(3)   \\
\midrule
$27$  & 17.23 & 18.02 & \textbf{18.09} \\
$108$ & 20.00 & \textbf{20.87} & 20.11 \\
$147$ & 20.81 & \textbf{21.13} & 20.66 \\
$300$ & 21.41 & \textbf{22.10} & 21.12 \\
\bottomrule
\end{tabular}
\caption{Mean PSNR on the test set for models with varying levels of equivariance.}
\label{tab:comparison_equivariance}
\end{table}

\section{Discussion and Conclusion}
We introduced rotation-equivariant spherical neural fields and used them to create RENI++, a natural illumination prior. Demonstrating how random samples from RENI++ always produce plausible illumination maps and RENI++'s usefulness for environment completion and inverse rendering. There are many exciting avenues for future research, for example, implementing RENI++ in larger inverse rendering pipelines where it could be a simple drop-in replacement for SH and using RENI++ for LDR to HDR image reconstruction. 

Our modifications in RENI++ have overcome two of the original limitations of the work, namely the \(O(n^{2})\) complexity of the Gram matrix and the reproduction of higher frequency details. However, some limitations remain. Because RENI++ has a prior for natural illumination, unlike SH, RENI++'s performance decreases when fit to indoor scenes. SH can also handle unnatural illuminations, such as a scene with multiple suns, something RENI++ will struggle to express.

Human vision has complex interactions between illumination, geometry and texture priors. For example, the Hollow Face Illusion \cite{hill_independent_1993} arises from face geometry priors overriding the lighting from above illumination prior; while in the Bas relief ambiguity \cite{belhumeur_bas-relief_1999}, geometric priors cause incorrect lighting estimation. Our model discounts these interactions, learning an illumination prior independently but not its interactions with other cues. 

We strictly allow only $SO(2)$ equivariance. However, considering typical camera coordinate systems, the up axis will sometimes not align with gravity when the camera is pointed up or down. For inverse problems, this would mean the gravity vector would need to be explicitly estimated (an accelerometer would resolve this). Alternatively, we could build our model with full $SO(3)$ equivariance but then learn a prior over the space of camera poses relative to gravity.

\section*{Acknowledgments}
James Gardner was supported by the EPSRC Centre for Doctoral Training in Intelligent Games \& Games Intelligence (IGGI) (EP/S022325/1). We would like to thank the attendees of Dagstuhl Seminar, 22121 - 3D Morphable Models and Beyond, for their valuable insights and discussions around this work.

%\vfill

\bibliographystyle{IEEEtran}
\bibliography{references}

\end{document}